\def\year{2022}\relax
\documentclass[letterpaper]{article} 
\usepackage{aaai22}  
\usepackage{times}  
\usepackage{helvet}  
\usepackage{courier}  
\usepackage[hyphens]{url}  
\usepackage{graphicx} 
\usepackage{natbib}  
\usepackage{caption} 
\DeclareCaptionStyle{ruled}{labelfont=normalfont,labelsep=colon,strut=off} 
\usepackage{algorithm}
\usepackage{algorithmic}

\newcommand{\ie}{\textit{i}.\textit{e}.}
\newcommand{\eg}{\textit{e}.\textit{g}.}
\usepackage{amsmath}
\usepackage{amssymb}
\usepackage{makecell}
\usepackage{multirow}
\usepackage{multicol}
\usepackage{booktabs}
\usepackage{pifont}

\usepackage{arydshln}
\usepackage[dvipsnames]{xcolor}
\DeclareMathOperator*{\argmax}{argmax}
\DeclareMathOperator*{\argmin}{argmin}

\usepackage{newfloat}
\usepackage{listings}
\floatstyle{ruled}
\newfloat{listing}{tb}{lst}{}
\floatname{listing}{Listing}
\title{Meta Input: How to Leverage Off-the-Shelf Deep Neural Networks}
\author{
	Minsu Kim\equalcontrib, Youngjoon Yu\equalcontrib, Sungjune Park\equalcontrib, Yong Man Ro\\
}
\affiliations{
	Image and Video Systems Lab, KAIST, South Korea\\
	
	\{ms.k, greatday, sungjune-p, ymro\}@kaist.ac.kr
}

\begin{document}

\maketitle

\begin{abstract}
	These days, although deep neural networks (DNNs) have achieved a noticeable progress in a wide range of research area, it lacks the adaptability to be employed in the real-world applications because of the environment discrepancy problem. Such a problem originates from the difference between training and testing environments, and it is widely known that it causes serious performance degradation, when a pretrained DNN model is applied to a new testing environment. Therefore, in this paper, we introduce a novel approach that allows end-users to exploit pretrained DNN models in their own testing environment without modifying the models. To this end, we present a \textit{meta input} which is an additional input transforming the distribution of testing data to be aligned with that of training data. The proposed meta input can be optimized with a small number of testing data only by considering the relation between testing input data and its output prediction. Also, it does not require any knowledge of the network's internal architecture and modification of its weight parameters. Then, the obtained meta input is added to testing data in order to shift the distribution of testing data to that of originally used training data. As a result, end-users can exploit well-trained models in their own testing environment which can differ from the training environment. We validate the effectiveness and versatility of the proposed meta input by showing the robustness against the environment discrepancy through the comprehensive experiments with various tasks.
\end{abstract}

\section{Introduction}
\begin{figure*}[t!]
	\begin{minipage}[b]{1.0\linewidth}
		\centering
		\centerline{\includegraphics[width=11.5cm]{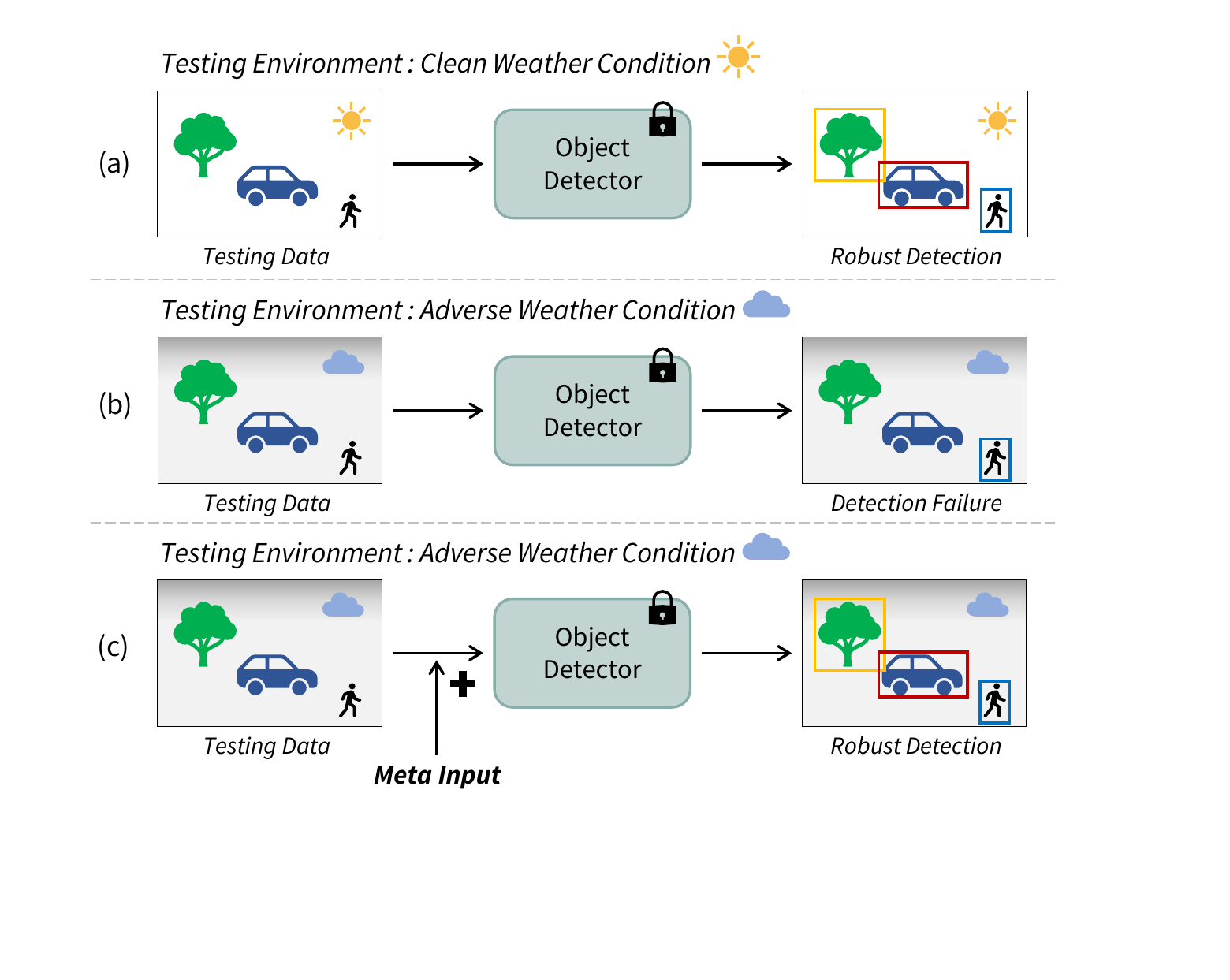}}
	\end{minipage}
	\caption{(a) describes that the object detector which is trained on clean weather data can perform the object detection successfully. However, (b) shows that, when end-users exploit the detector under the adverse weather condition, it usually fails the detection because of the mismatch between the training and testing environments. To alleviate the problem, we introduce a meta input which is embedded into testing data and transforms the distribution of testing data (\ie, adverse weather) to that of training data (\ie, clean weather). So that, the detector becomes to conduct the object detection properly.}
	\label{fig:1}
\end{figure*}

Recently, as the deep learning has achieved a great development, deep neural networks (DNNs) have shown remarkable performances in various research areas, such as computer vision \cite{he2016resnet}, natural language processing \cite{vaswani2017attention}, and speech processing \cite{amodei2016deepspeech2}. Nevertheless, there exists one significant problem to be solved in utilizing DNNs robustly in the real-world, that is, the environment discrepancy problem. The environment discrepancy problem occurs when the training data distribution and testing data distribution are mismatched, and it results in the serious performance degradations of DNNs \cite{klejch2019SA_meta,touvron2019TraintTestResolutionDis,teney2020OOD}. Therefore, although end-users want to exploit well-trained DNNs in their own testing environment, they would fail to experience the powerfulness of DNNs because of the aforementioned problem. For example, as described in Fig. \ref{fig:1}(a), an end-user tries to adopt an object detection model which is trained with clean weather data, and of course, it detects objects successfully under thte same clean weather condition. However, as shown in Fig. \ref{fig:2}(b), when the user wants to detect objects under adverse weather condition, the detection model would fail to conduct the robust object detection, because the testing environment (\ie, adverse weather) differs from the training environment (\ie, clean weather) \cite{huang2022sfa,foggy}. Therefore, end-users are recommended to find and exploit the DNN models well-trained on the training data that is consistent with their own testing environment. One possible approach to alleviate such a problem is Domain Adaptation (DA) which aims to reduce the domain gap between the source and target domains by learning the domain invariant representations \cite{xiao2021DA1,saito2018DA2,bousmalis2017DA3,ganin2015DA4,long2017DA5}. However, such DA methods are usually required to know the internal architecture of the network and have an access to both source and target data simultaneously for learning the domain invariant features. Therefore, it is time-consuming and difficult for end-users to understand the behavior of the network and use both kinds of data.

In this paper, we focus on how to make end-users enjoy the beneficial performances of well-trained DNN models even with different testing data distribution. To this end, we introduce a method that end-users can adapt the pretrained model into their testing environment without any knowledge of model architecture or finetuning the model, by using only a small number of data in testing environment. Motivated by the recent success in the input-level transformation to convert the originally learned task to another task \cite{elsayed2018AR1}, instead of modifying the weight parameters of the pretrained models, we propose to use an additional input, called \textit{meta input}, to match the distributions of testing data with that of training data. Specifically, we suppose that an end-user wants to adopt a pretrained model under different testing environment with a few labeled/unlabeled testing data only, while the user cannot have an access to the training data which is used to pretrain the model. Then, the proposed meta input can be optimized to transform the testing data distribution to be aligned with the training data distribution where the pretrained model operates properly. After that, the meta input can be embedded into the testing data to make the pretrained model perform well under different testing environment. For example, as shown in Fig. \ref{fig:1}(c), the meta input is embedded into the testing data, so that the pretrained detection model conducts the robust object detection even under adverse weather condition without modifying its weight parameters.

The proposed meta input can be optimized simply with any gradient-based training algorithm by using a few labeled, or unlabeled, data of testing environment. With the meta input, the learned knowledge of pretrained DNN models can be extended to diverse testing environments without knowing the network architecture and modifying its weight parameters. Therefore, end-users can enjoy the powerfulness of off-the-shelf DNNs on their own testing environment. We verify both effectiveness and practicality of the proposed meta input in the real-world through the extensive experiments in the three tasks, image classification, object detection, and visual speech recognition.

Our contributions can be summarized as follows:
\begin{itemize}
	\item Since the proposed meta input can match the distribution of testing data with that of training data, the knowledge the pretrained DNN models already learned can be utilized properly even under different environments.
	\item Different from the existing DA methods, the proposed method does not require any knowledge of the model architecture, modification of its weight parameters and training data (which is used for pretraining the model), and it only needs a small number of testing data.
	\item The effectiveness and versatility of the proposed meta input are corroborated by the comprehensive experiments on three practical tasks, image classification, object detection, and visual speech recognition.
\end{itemize}

\section{Related Work}
\subsection{Domain Adaptation}
Deep Neural Networks (DNNs) have been widely adopted to extract the generalized feature representation of the data. To train such a generalized DNNs, it assumes that both training and testing data are originated from the same distribution and share some similar joint probability distribution. In the real-world scenario, however, this constraint is easily violated, because each training and test data can be drawn from different distributions. To tackle the aforementioned problems, researchers have devoted their efforts on a research field called Domain Adaptation (DA) \cite{wang2018deep,kang2019contrastive}. DA is a technique that enables DNNs learned with sufficient label and data size (\ie, source domain) to perform and generalize well on data sampled from different distributions (\ie, target domain). DA can be categorized into discrepancy-based methods \cite{long2015learning, long2017DA5, zhang2015deep}, adversarial-based methods \cite{bousmalis2017DA3,ganin2015DA4}, and reconstruction-based methods \cite{glorot2011domain, ghifary2016deep}. Most of the existing works of DA focus on enhancing the model performance by adopting an additional cost of the model architecture modification or re-training. Moreover, they usually need both source and target domain data simultaneously. Different from DA, we propose a novel method called \textit{meta input} which does not require knowing about the model architecture and finetuning of the model. The proposed meta input is an additional input which can be obtained by using a small number of testing (\ie, target) data only without any training (\ie, source) data. Then, the obtained meta input can improve the performances on testing environment by transforming the testing input distribution into the training data distribution.

\subsection{Input Transformation}
Recently, input transformation methods attract large attention \cite{elsayed2018AR1, chen2021AR2, dinh2022AR3, zheng2021AR4, neekhara2022AR5} with its potential to interact with a trained model without modification of its weight parameters. For example, DNNs trained to classify classes of samples from a source task (\eg, ImageNet classification) can be reprogrammed to classify hand-written digits target task (\eg, digits classification)~\cite{elsayed2018AR1}. To this end, a mapping function between the class labels of the source task and the class labels of the target task should be organized in advance. Once such a class mapping process is finished, the frame-shaped adversarial perturbation is applied surrounding the input image to perform the target task. In this paper, we try to provide a framework using input transformation, so that end-users can work on with an off-the-shelf DNNs developed for a certain task without considering the gap between training and testing environments. Different from the aforementioned input transformation works, we do not consider the different tasks scenario but focus on how to use the pretrained off-the-shelf DNNs in end-users' environments that are usually distinct from the development environment.

\begin{figure*}[t!]
	\begin{minipage}[b]{1.0\linewidth}
		\centering
		\centerline{\includegraphics[width=11.0cm]{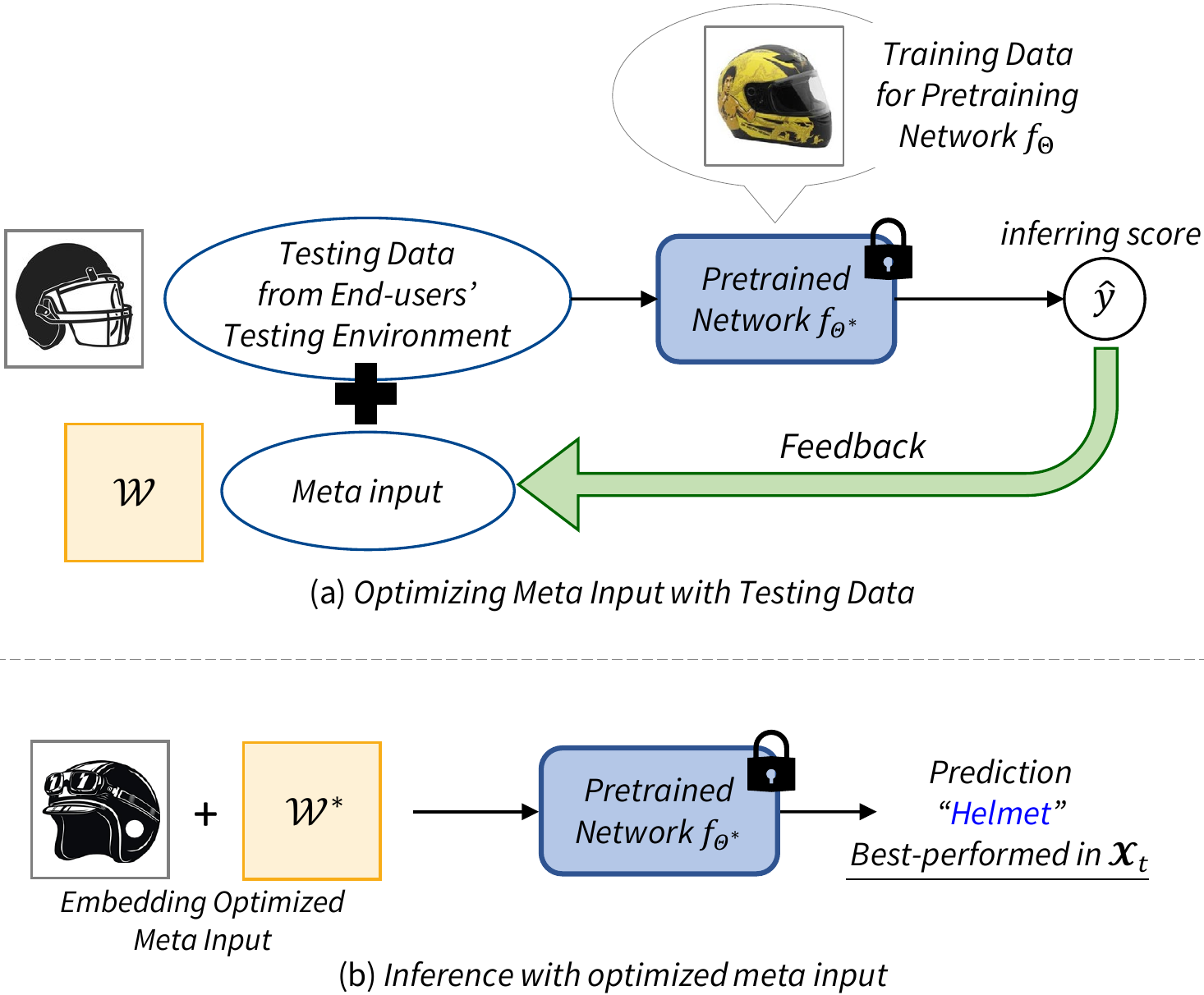}}
	\end{minipage}
	\caption{Overview of the proposed meta input. When the target testing data has different distribution from the source training data which is used for pretraining the network, (a) shows that a meta input can be optimized by investigating the input and output relationships. During this meta input optimization, the weight parameters $\Theta^*$ are not modified at all. After optimization, (b) describes that a meta input is added into the testing samples that a user wants to work on. By embedding a meta input, the distribution of testing data becomes aligned with that of training data, so that the network can make a robust prediction.}
	\label{fig:2}
\end{figure*}

\section{Proposed Method}
\subsection{Problem Formulation}
In this paper, we consider that an end-user has an access to a pretrained DNN which can perform a specific task (\eg, image classification, \textit{etc.}). The given neural network $f$ is pretrained on the training data, called source domain data, $\boldsymbol{\mathcal{S}}=\{\boldsymbol{x}^i_s, \boldsymbol{y}^i_s\}^{N_s}_{i=1}\in\{\boldsymbol{\mathcal{X}_s},\boldsymbol{\mathcal{Y}_s}\}$ consisting of $N_s$ samples, as follows,
\begin{align}
\Theta^* = \argmin_\Theta \mathbb{E}_i\left[\mathcal{L}(f_{\Theta}(\boldsymbol{x}^i_s), \boldsymbol{y}^i_s)\right],
\label{eq:1}
\end{align}
where $\boldsymbol{x}^i_s$ and $\boldsymbol{y}^i_s$ are the $i$-th source domain sample and its corresponding label, respectively, $\mathcal{L}(\cdot)$ represents the objective function defined for the task, and $\Theta$ is learnable parameters of the neural network. Then, the pretrained model can be regarded as a mapping function $f_{\Theta^*}: \boldsymbol{\mathcal{X}_s} \rightarrow \boldsymbol{\mathcal{Y}_s}$ performing predictions on the source domain data properly, which is parameterized by $\Theta^*$. 

Let us assume that an end-user wants to apply the model on their own testing data, called target domain data, $\boldsymbol{\mathcal{T}}=\{\{\boldsymbol{x}^i_{l,t},  \boldsymbol{y}^i_{l,t}\}^{N_{l,t}}_{i=1},\{\boldsymbol{x}^j_{u,t}\}_{j=1}^{N_{u,t}}\}$ consisting of $N_{l,t}$ labeled samples and $N_{u,t}$ unlabeled samples, where $\{\boldsymbol{x}_{l,t}, \boldsymbol{x}_{u,t}\} \in \boldsymbol{\mathcal{X}_t}$ and $\boldsymbol{y}_{l,t} \in \boldsymbol{\mathcal{Y}_t}$. The labeled samples are typically very few compared to the source domain data, \ie, $0 \leq N_{l,t} \ll N_s$. Usually, target domain can be changed dynamically depending on which environment a user wants to apply, so that the distribution of target domain data can differ from that of source domain data. In this case, the pretrained model would not make predictions properly on the data that a user wants to test (\eg, $f_{\Theta^*} : \boldsymbol{\mathcal{X}_t} \nrightarrow \boldsymbol{\mathcal{Y}_t}$), because of the mismatch of training and testing environments. To deal with it, in this paper, we present a meta input $\mathcal{W}$ which is an additional input that will be applied into the testing data to make the pretrained model perform predictions properly on target domain data as well. Therefore, we aim to construct a mapping function $f_{\Theta^*}: (\boldsymbol{\mathcal{X}_t} + \mathcal{W}) \rightarrow \boldsymbol{\mathcal{Y}_t}$ without modifying the originally learned weight parameters $\Theta^*$, by only adding the proposed meta input $\mathcal{W}$ to the testing data sampled from $\boldsymbol{\mathcal{X}_t}$.

\subsection{Meta Input}
The proposed meta input $\mathcal{W}$ is learnable and universal, so that it can be embedded into every target domain data once after it is optimized. The overall optimization flow is described in Fig. \ref{fig:2}(a). The meta input can be optimized by investigating the relationships between the target inputs covered by the meta input and the model prediction scores. Then, the optimized meta input can convert the distribution of target domain data into that of training source domain data, and it makes the model perform properly on target domain data as well. For the better understanding, we suppose the data type as the image in the remaining paper. When the target input image comes in, a meta input $\mathcal{W} \in \mathbb{R}^{H \times W \times C}$ occupies the entire image, having the same size with the input images, where $H$, $W$, and $C$ are the dimensions of height, width, and channel, respectively. 

To transform the target domain images into the source domain in the latent space, we apply input-level transformation by adding the meta input $\mathcal{W}$ to the target domain images as follows, $\tilde{\boldsymbol{x}}_t = \boldsymbol{x}_t + \mathcal{W}$, where $\tilde{\boldsymbol{x}}_t$ is the transformed image. The optimal meta input $\mathcal{W}^*$ can be acquired by solving the following optimization problem,
\begin{align}
\mathcal{W}^* = \argmin_{\mathcal{W}} \mathbb{E}_i\left[\mathcal{L}(f_{\Theta^*}(\tilde{\boldsymbol{x}}_{l,t}^i),\boldsymbol{y}^i_{l,t})\right],
\label{eq:2}
\end{align}
where it can be solved using gradient-based training algorithms. By minimizing the task loss $\mathcal{L}(\cdot)$ without updating the trained model parameter $\Theta^*$, the meta input can be optimized, so that the target domain images have a similar distribution to the source domain data in the latent space. The optimized meta input $\mathcal{W}^*$ can be applied into target input image via element-wise addition, as shown in Fig. \ref{fig:2}(b). Therefore, with the optimized meta input, the given pretrained model $f_{\Theta^*}$ can robustly operate on the testing data with the following formula, $\hat{\boldsymbol{y}}=f_{\Theta^*}(\boldsymbol{x}_{u,t} + \mathcal{W}^*)$. Here, $\hat{\boldsymbol{y}}$ is the prediction result for the unlabeled target data $\boldsymbol{x}_{u,t}$ that a user wants to test. 
Even if we assume that the labeled target domain data $\boldsymbol{x}_{l,t}$ is available (\ie, $N_{l,t} > 0$), the existing unsupervised methods \cite{lee2013selftraining1,xie2020selftraining3,tzeng2017adversarial,ganin2015DA4} can be adopted for solving the optimization problem, when there is no labeled target domain data (\ie, $N_{l,t} = 0$). For example, we can employ one of self-training methods \cite{lee2013selftraining1, xie2020selftraining3} to firstly perform pseudo labeling on the unlabeled samples $\boldsymbol{x}_{u,t}$. And then, the obtained pseudo labels $\bar{\boldsymbol{y}}_{u,t}$ are used for training a meta input. The model confidence-based pseudo labeling can be written as follows,
\begin{align}
\bar{\boldsymbol{y}}_{u,t} = \argmax f_{\Theta^*}(\boldsymbol{x}_{u,t}), \quad \text{if} \,\,\, p(\bar{\boldsymbol{y}}_{u,t}|\boldsymbol{x}_{u,t}) > \alpha
\end{align}
where $\alpha$ determines the model confidence range for using predictions as pseudo labels. With the obtained pseudo labels $\bar{\boldsymbol{y}}_{u,t}$, we can easily solve the optimization in Eq. (\ref{eq:2}).

In the following sections, we will show the effectiveness of the meta input with comprehensive experiments and corroborate its versatility with diverse applications including image classification, object detection, and visual speech recognition.

\section{Experiments}
We verify the effectiveness of the proposed meta input in three tasks (\ie, image classification, object detection, and visual speech recognition) to show its versatility and practicality. Image classification experiments can be divided into two scenarios: 1) domain shifted environment and 2) noisy environment. Basically, we perform the experiments in a supervised setting that supposes a small number of the target dataset are available for optimizing the meta input. Then, we extend the experiments to show the meta input can be learned in an unsupervised way.

\subsection{Domain Shifted Image Classification}
We perform domain shifting experiments that target testing data differs from the source training data.

\subsubsection{Datasets \& Setup} We evaluate the proposed meta input in digits classification using MNIST \cite{lecun1998MNIST}, USPS \cite{hastie2009USPS}, and SVHN \cite{netzer2011SVHN}. The input images are resized into $28\times28$ and converted into grayscale. Each dataset contains 10 classes of digits, and three domain shifted environments are set as similar with \cite{tzeng2017adversarial}: MNIST $\rightarrow$ USPS, USPS $\rightarrow$ MNIST, and SVHN $\rightarrow$ MNIST. Please note that each source dataset (\textit{left}) is used to build the pretrained off-the-shelf DNN models. The evaluation is conducted with a test set of target dataset by using classification accuracy (\%) as evaluation metric. We use a simple 3-layered CNN followed by fully connected layers for the classification. The meta input is optimized by using different amounts (1\%, 30\%, 70\%, and 100\%) of training data from the target dataset. Please note the network parameters are not modified during optimization of meta input.

\begin{table}[t!]
	\renewcommand{\arraystretch}{1.3}
	\renewcommand{\tabcolsep}{3mm}
	\centering
	\resizebox{0.87\linewidth}{!}{
		\begin{tabular}{cccc}
			\Xhline{3\arrayrulewidth}
			\makecell{\textbf{Ratio of} \\ \textbf{target data}} & \makecell{\textbf{MNIST} \\ \textbf{$\rightarrow$ USPS}} & \makecell{\textbf{USPS} \\ \textbf{$\rightarrow$ MNIST}} & \makecell{\textbf{SVHN} \\ \textbf{$\rightarrow$ MNIST}}\\ \hline
			Baseline & 81.73 & 34.33 & 60.64 \\ \hdashline
			\textbf{1\%} & 88.28 & 80.47 & 83.59 \\
			\textbf{30\%} & 95.31 & 89.58 & 92.97 \\
			\textbf{70\%} & 96.09 & 89.45 & 94.01 \\
			\textbf{100\%} & 96.48 & 90.23 & 94.53 \\
			\Xhline{3\arrayrulewidth}
	\end{tabular}}
	\caption{Domain shift experiments on digits datasets, MNIST, USPS, and SVHN. The evaluation metric is classification accuracy (\%).}
	\label{table:1}
\end{table}

\begin{table}[!t]
	\renewcommand{\arraystretch}{1.2}
	\centering
	\resizebox{1.0\linewidth}{!}{
		\begin{tabular}{ccccccc}
			\Xhline{3\arrayrulewidth}
			\multirow{2}{*}{\begin{tabular}[c]{@{}c@{}}\textbf{Ratio of}\\ \textbf{target data}\end{tabular}} & \multicolumn{3}{c}{\textbf{CIFAR10}} & \multicolumn{3}{c}{\textbf{CIFAR100}} \\ \cmidrule(lr){2-4} \cmidrule(lr){5-7} 
			& 33 dB & 26 dB & 23 dB & 33 dB & 26 dB & 23 dB \\ \hline
			Baseline & 85.89 & 60.79 & 38.05 & 56.60 & 32.08  & 19.82 \\ \hdashline
			\textbf{1\%} & 88.41 & 75.46 & 63.35 & 60.16 & 41.67 & 29.69 \\
			\textbf{30\%} & 89.06 & 76.92 & 66.08 & 60.61 & 42.77 & 30.08 \\
			\textbf{70\%} & 89.19 & 77.34 & 66.47 & 60.87 & 44.27 & 31.71 \\
			\textbf{100\%} & 89.45 & 78.06 & 66.73 & 61.13 & 44.34 & 32.81 \\
			\Xhline{3\arrayrulewidth}
	\end{tabular}}
	\caption{Noise robust classification under different Gaussian Noise PSNR (dB) and ratio of data. The evaluation metric is classification accuracy (\%).}
	\label{table:2}
\end{table}

\subsubsection{Results} Table \ref{table:1} shows results of domain shift experiments on digit datasets. The \textit{baseline} performances refer to the performances on the target data obtained by the model pretrained on the source data without a meta input. As the table shows, when the model is directly applied to different domain data, the performances are much lower compared to when it is applied to the same domain data (MNIST $\rightarrow$ MNIST: 99.61\%, USPS $\rightarrow$ USPS: 98.31\%, and SVHN $\rightarrow$ SVHN: 91.54\%). In contrast, when the proposed meta input is optimized by using only 1\% of target domain data, the performances of all cases are significantly improved. For example, the classification accuracy is improved by about 46\% in USPS $\rightarrow$ MNIST, which is about 134\% relative improvement from the \textit{Baseline}. Moreover, when the users can access more data of the target domain, they can experience more effective performances without modifying the model but just learning the meta input. For instance, when 30\% of target data is available, the performances are improved by 13.6\%, 55.3\%, and 32.3\% on MNIST $\rightarrow$ USPS, USPS $\rightarrow$ MNIST, and SVHN $\rightarrow$ MNIST, respectively.

\subsection{Noisy Image Classification}
We also conduct noisy image classification experiments assuming that the image can be corrupted with unseen noise which leads to the performance drop. In this case, the meta input can be utilized to help the model to perform noise robust classification.

\begin{table}[!t]
	\renewcommand{\arraystretch}{1.3}
	\centering
	\resizebox{0.87\linewidth}{!}{
		\begin{tabular}{cccccc}
			\Xhline{3\arrayrulewidth}
			\textbf{Dataset} & \textbf{Method} & \textbf{GN} & \textbf{GB} & \textbf{SP} & \textbf{SN} \\ \hline
			\multirow{2}{*}{CIFAR10} & Baseline & 60.79 & 53.36 & 40.81 & 38.68 \\ \cdashline{2-6}
			&\textbf{Ours} & 76.92 & 84.24 & 56.51 & 52.60  \\ \hline
			\multirow{2}{*}{CIFAR100} & Baseline & 32.08 & 32.43 & 17.38 & 18.13 \\ \cdashline{2-6}
			&\textbf{Ours} & 42.77 & 50.39 & 24.35 & 27.08  \\
			\Xhline{3\arrayrulewidth}
	\end{tabular}}
	\caption{Classification results under different noise types (GN: Gaussian Noise, GB: Gaussian Blur, SP: Salt and Pepper, SN: Speckle Noise) on CIFAR10 and CIFAR100. The meta input is optimized on 30\% of target data. The evaluation metric is classification accuracy (\%).}
	\label{table:3}
\end{table}

\begin{table}[!t]
	\renewcommand{\arraystretch}{1.3}
	\renewcommand{\tabcolsep}{3.5mm}
	\centering
	\resizebox{0.66\linewidth}{!}{
		\begin{tabular}{ccc}
			\Xhline{3\arrayrulewidth}
			\textbf{Method} & \textbf{CIFAR10} & \textbf{CIFAR100} \\ \hline
			Baseline & 45.49 & 23.51 \\ \hdashline
			\textbf{Ours} & 69.79 & 38.35 \\ \Xhline{3\arrayrulewidth}
	\end{tabular}}
	\caption{Classification results under four comprehensive noise types (GN, GB, SP, and SN) on CIFAR10 and CIFAR100 by using 30\% of target data. The evaluation metric is classification accuracy (\%).}
	\label{table:4}
\end{table}

\subsubsection{Datasets \& Setup} We evaluate the proposed meta input in noisy image classification using CIFAR10 and CIFAR100 \cite{krizhevsky2009learning} under noise corrupted environment. CIFAR10 consists of 10 classes of objects and CIFAR100 is comprised of 100 classes. Both datasets have images size of $32 \times 32$. The classification accuracy (\%) is used as evaluation metric.
We use VGG-16 \cite{simonyan2014very}, and pretrain the network with the clean image data (\ie, source data). And then, the meta input is optimized by using different amounts (1\%, 30\%, 70\% and 100\%) of training set from noise corrupted images (\ie, target data). We measure the performances with various noise types, including Gaussian Noise (GN), Gaussian Blur (GB), Salt and Pepper (SP), and Speckle Noise (SN). For GN, we vary the range of average Peak Signal to Noise Ratio (PSNR) from 33 dB to 23 dB.

\subsubsection{Results} Table \ref{table:2} shows experimental results of noisy image classification using GN with varying PSNR. The \textit{baseline} is the pretrained network achieving 92.1\% and 70.6\% accuracies on the clean CIFAR10 and CIFAR100, respectively. However, as shown in the table, the \textit{baseline} shows performance drops with GN which is unseen during the network training. With the proposed meta input which is optimized by using only 1\% noisy images from target data, the classification accuracy is significantly enhanced. For example, the accuracy is improved by about 25\% on CIFAR10 under GN at 23 dB. It demonstrates the effectiveness of the meta input when the testing image is corrupted and not consistent with model training environment.

Table \ref{table:3} describes the classification results under different types of noise. In this experiment, we use only 30\% of target domain data to optimize the meta input. Compared to the \textit{baseline} the meta input consistently improves the performances regardless of noise types on both datasets. Finally, as shown in Table \ref{table:4}, we compose the target domain data by randomly applying four types of noise on the images (\ie, comprehensive noise situation). As shown in the table, our method improves the performances by 24.3\% and 14.8\% on each dataset. The experimental results also corroborate that the meta input is effective against noisy image classification by extending the learned knowledge of the models on the clean images to noisy testing environments.

\begin{table}[!t]
	\renewcommand{\arraystretch}{1.3}
	\renewcommand{\tabcolsep}{4.0mm}
	\centering
	\resizebox{0.75\linewidth}{!}{
		\begin{tabular}{ccc}
			\Xhline{3\arrayrulewidth}
			\makecell{\textbf{Ratio of} \\ \textbf{target data}} & \makecell{\textbf{VOC} \\ \textbf{$\rightarrow$ Comic2k}} & \makecell{\textbf{Cityscapes} \\ \textbf{$\rightarrow$ Foggy}} \\ \hline
			Baseline  & 21.7 & 23.5 \\ \hdashline
			\textbf{10\%} & 25.6 &  25.5 \\
			\textbf{30\%} & 26.0 &  25.8 \\
			\textbf{70\%} & 27.1 &  26.1 \\
			\textbf{100\%} & 27.4 & 26.4 \\
			\Xhline{3\arrayrulewidth}
	\end{tabular}}
	\caption{Object detection performances (mAP).}
	\label{table:5}
	\vspace{-0.2cm}
\end{table}

\subsection{Object Detection}
The object detection has also shown degraded detection performances on different testing environment. Therefore, in this section, we show the effectiveness of the proposed meta input on the object detection.

\subsubsection{Datasets \& Setup} We verify the effectiveness of a meta input in two scenarios: 1) real to synthetic images and 2) clear to foggy weather conditions. For the first scenario, PASCAL VOC dataset \cite{pascal} is chosen for the real data to pretrain the off-the-shelf object detector. We compose source training data with VOC2007 and VOC2012-trainval sets having total 16,551 image samples. Also, Comic2k \cite{comic} is used for the synthetic dataset having total 2k of images (1k for a train set and other 1k for a test set). Please note that, we use 1k train set of Comic2k for training a meta input and another 1k test set for its evaluation. For the second scenario, we use Cityscapes \cite{cityscapes} and Foggy Cityscapes \cite{foggy} datasets as for each clear and foggy weather condition, respectively. Cityscapes dataset is used for pretraining the detector, and we use the train set of Foggy Cityscapes dataset for training a meta input, while not modifying the weight parameters of the detector. The evaluation metric is mean average precision (mAP).
In our experiments, we employ SSD300 \cite{ssd} which is built upon VGG-16 architecture \cite{vgg} taking input size of $300 \times 300$. We use object detection loss for its classification and regression, following \cite{ssd}. After that, we train a meta input with various amounts of the train set from each target domain data, Comic2k and Foggy Cityscapes (10\%, 30\%, 70\%, and 100\%).

\subsubsection{Results} The experimental results are mainly shown in Table \ref{table:5} for both scenarios: real to synthetic (VOC $\rightarrow$ Comic2k) and clear to foggy (Cityscapes $\rightarrow$ Foggy). The leftmost column represents the ratio of target data which is used for training a meta input. In the first scenario, the \textit{baseline} originally achieves $77.1$ mAP on VOC source dataset, however, the detection performance decreases to $21.7$ mAP, when it is evaluated on Comic2k target dataset. Compared to the \textit{baseline}, a meta input optimized by using only 10\% of target data can bring about 18\% of relative performance improvement. In the second scenario, the detection performance of the \textit{baseline} is degraded to $23.5$ mAP on Foggy Cityscapes target dataset, while it acquires $35.6$ mAP on Cityscapes source dataset. Also, in similar with the first scenario, we can obtain performance improvement by utilizing the proposed meta input. The experiments corroborate the effectiveness of a meta input on various situation (real $\rightarrow$ synthetic and clean $\rightarrow$ foggy) via aligning the distribution of target testing environment with that of source training environment.

\begin{table}[t!]
	\renewcommand{\arraystretch}{1.3}
	\renewcommand{\tabcolsep}{3.0mm}
	\centering
	\resizebox{0.9\linewidth}{!}{
		\begin{tabular}{cccccc}
			\Xhline{3\arrayrulewidth}
			Method & \textbf{S\#1} & \textbf{S\#2} & \textbf{S\#20} & \textbf{S\#22} & \textbf{Mean} \\ \hline
			Baseline & 17.04 & 9.02 & 10.33 & 8.13 & 11.12 \\ \hdashline
			\textbf{1 minute} & 10.71 & 3.77 & 6.10 & 4.20 & 6.20 \\
			\textbf{3 minute} & 9.53 & 3.21 & 5.83 & 3.77 & 5.58 \\
			\Xhline{3\arrayrulewidth}
	\end{tabular}}
	\caption{Speaker adaptation experiments on GRID dataset. `K minute' means the length of each target data used to train our method. The evaluation metric is word error rate (WER, \%), and `\textbf{S\#K}' means `Subject K' in the dataset.}
	\label{table:6}
\end{table}

\subsection{Visual Speech Recognition} Many existing visual speech recognition methods show a performance drop, when it is applied to unseen speakers during training \cite{abdel2013speakeradapt}. In this experiment, we show that meta input can be adopted for speaker adaptation \cite{kim2022speaker}, and the speech recognition performance for unseen speakers can be improved.

\subsubsection{Datasets \& Setup} We evaluate the proposed meta input in visual speech recognition task using GRID \cite{cooke2006grid} dataset. It contains utterances from 33 speakers. We use the unseen-speaker setting of \cite{assael2016lipnet} where subjects 1, 2, 20, and 22 are used for evaluation and the remainders are used for training. We split the evaluation data of each subject into two sets, one for training the meta input and the other for testing. Mouth region is cropped from frames, and resized into $64\times128$. All frames are consistently added with one meta input frame. For training the meta input, we sample data of 1 minute and 3 minute lengths from the first set, and examine the effect of meta input under different amounts of data.
For the network, we use modified version of LipNet \cite{assael2016lipnet}, one of the popular visual speech recognition architecture, and word-level CTC loss \cite{graves2006ctc} is employed for training. The network is composed of three 3D convolutions, two 2D convolutions, and a two-layered Bi-GRU. The model is trained with the data of all subjects except the test subjects (1, 2, 20, and 22), then the meta input is optimized with 1 minute and 3 minute lengths of data of each test subject. Word Error Rate (WER) is reported for the performance measurment.

\subsubsection{Results} The speaker adaptation results using the proposed meta input are shown in Table \ref{table:6}. When the model is directly applied to unseen speakers, it achieves 17.04\%, 9.02\%, 10.33\%, and 8.13\% WER for subject 1, 2, 20, and 22, respectively, and 11.12\% mean WER, while mean WER on seen speaker is 5.6\% \cite{assael2016lipnet}. It is because the unseen speaker characteristics (\eg, lip appearances) are not seen during the model training process. By embedding the meta input optimized with just 1 minute lengths of target data, the error rate decreases significantly. The performances are improved by about 6.3\%, 5.3\%, 4.2\%, and 3.9\% WER on each subject from the \textit{baseline}, which is a large improvement of 4.92\% mean WER. The meta input pushes unseen visual representations into the learned visual distributions, so that the lip appearances and movements of unseen speakers can be properly modelled via the pretrained model. Also, when we increase the amount of target data to 3 minute lengths, it enhances the performances more by achieving 5.58\% mean WER. Please note that, we do not modify the weight parameters of the pretrained model, and instead, we just transform its input with the addable meta input to fully utilize the learned model knowledge.

\begin{table}[]
	\renewcommand{\arraystretch}{1.3}
	\centering
	\resizebox{1.0\linewidth}{!}{
		\begin{tabular}{cccccc}
			\Xhline{3\arrayrulewidth}
			\textbf{Method} & \textbf{S\#1} & \textbf{S\#2} & \textbf{S\#20} & \textbf{S\#22} & \textbf{Mean} \\ \hline
			\citet{assael2016lipnet} & 17.0 & 9.3 & 10.3 & 7.8 & 11.1 \\
			\citet{yang2020TVSR} & - & - & - & - & 9.1 \\
			\citet{zhang2021DVSR} & - & - & - & - & 7.8 \\ \hline
			\textbf{Ours} & \textbf{13.9} & \textbf{4.6} & \textbf{7.7} & \textbf{4.7} & \textbf{7.7} \\
			\Xhline{3\arrayrulewidth}
	\end{tabular}}
	\caption{Unsupervised speaker adaptation experiments on GRID dataset. The evaluation metric is word error rate (WER, \%), and `\textbf{S\#K}' means `Subject K' in the dataset.}
	\label{table:7}
	\vspace{-0.2cm}
\end{table}

\subsection{Unsupervised Example}
We have shown the effectiveness of the meta input in the supervised training setting by using a small number of data from the target testing domain. However, since the meta input is decoupled with the network internal architecture, it can be optimized via any advanced learning algorithms. Therefore, in this section, we validate whether the meta input can be obtained in the unsupervised manner with visual speech recognition. We use all data of the unseen speakers 1, 2, 20, and 22 without their labels, following \cite{assael2016lipnet}, and we adopt a simple self-training method acquiring pseudo labels from the pretrained model. We use the predictions of the pretrained model which confidence scores are higher than 0.9. Then, the meta input is optimized with the obtained pseudo-labels, as it does in the supervised setting. Table \ref{table:7} shows the speaker adaptation results with the unsupervised learning method. As shown in the table, compared to LipNet \cite{assael2016lipnet}, our meta input method achieves more robust visual speech recognition performances by improving about 3.4\% mean WER. Moreover, we compare our method with state-of-the-art visual speech recognition methods \cite{yang2020TVSR, zhang2021DVSR}. Since the previous works did not report the performances per subject, we compare mean WER only. Compared with them, our method outperforms the existing methods \cite{yang2020TVSR, zhang2021DVSR}, even with a much simpler architecture \cite{assael2016lipnet}. Such results also support that the proposed meta input can be incorporated with diverse learning methods and it is effective even if the annotations are not available.

\begin{table}[!t]
	\renewcommand{\arraystretch}{1.3}
	\centering
	\resizebox{0.91\linewidth}{!}{
		\begin{tabular}{cccccc}
			\Xhline{3\arrayrulewidth}
			& \multicolumn{5}{c}{\textbf{Accuracy (\%)}} \\ \cline{2-6}
			Ratio (\%) & 0 & 1 & 30 & 70 & 100 \\ \hline
			Baseline & 34.33 & - & - & - &- \\ 
			Batch Normalization & - & 47.79 & 48.31 & 48.31 & 48.31 \\ \hdashline
			\textbf{Ours} & \bf - & \bf 80.47 & \bf 89.58 & \bf 89.45 & \bf 90.23 \\ \Xhline{3\arrayrulewidth}
	\end{tabular}}
	\caption{Comparison with Batch Normalization adaptation on Image Classification (USPS $\rightarrow$ MNIST). `Ratio' means the percentage of target data used for training batch normalization adaptation and the proposed meta input.}
	\label{table:8}
\end{table}

\begin{table}[!t]
	\renewcommand{\arraystretch}{1.3}
	\renewcommand{\tabcolsep}{3.5mm}
	\centering
	\resizebox{0.8\linewidth}{!}{
		\begin{tabular}{cccc}
			\Xhline{3\arrayrulewidth}
			& \multicolumn{3}{c}{\textbf{mean WER (\%)}} \\ \cline{2-4}
			Length (minute)      & 0        & 1         & 3 \\ \hline
			Baseline             & 11.12    & -         & - \\ 
			Batch Normalization  & -        & 12.27     & 12.37 \\ \hdashline
			\textbf{Ours}        & \bf -    & \bf 6.20  & \bf 5.58 \\ \Xhline{3\arrayrulewidth}
	\end{tabular}}
	\caption{Comparison with Batch Normalization adaptation on Visual Speech Recognition (GRID dataset). `Length' means the duration of target data used for training batch normalization adaptation and the proposed meta input.}
	\label{table:9}
	\vspace{-0.3cm}
\end{table}

\subsection{Discussion}
In this section, we discuss the comparison between the proposed meta input and the batch normalization adaptation \cite{li2018adabn} which can be finetuned with target data while fixing the weight parameters of the network. Table \ref{table:8} shows the experimental results on domain shifted image classification (USPS $\rightarrow$ MNIST). As shown in the table, the performance is improved by adapting the batch normalization only to the target data. However, we find that compared to the proposed meta input, the performance gain by using batch normalization adaptation is much lower. Furthermore, we also conduct the comparison experiment on visual speech recognition with GRID dataset, as described in Table \ref{table:9}. In this experiment, we observe that the batch normalization adaptation even shows degraded performance compared with the baseline. However, the proposed meta input enhances the performances consistently. For example, only adapting batch normalization degrades the performance by 1.15\% WER while only optimizing meta input increases the performance by 4.92\% WER, for the 1 minute adaptation case. From these results, we can confirm that the proposed meta input is effective regardless of the difficulty of tasks, and the network architecture.

\section{Conclusion}
We have proposed a novel method, called \textit{meta input}, to fully utilize the pretrained off-the-shelf DNN models by alleviating the mismatch between training and testing environments. The proposed meta input can transform the testing data to have similar distribution with the training data by being added to the testing input simply. Since it is not subject to the network internal architecture, it can be optimized by any advanced learning algorithms. Therefore, with the meta input, end-users can fully enjoy the beneficial advantages of off-the-shelf DNN models. The effectiveness and versatility of the meta input is comprehensively validated with various applications of image classification, object detection, and visual speech recognition. We hope that the meta input can contribute to advance the date that beneficial DNN models can be widely applied into in the real-world.

\bibliography{aaai22}

\end{document}